\documentclass[conference]{IEEEtran}
\IEEEoverridecommandlockouts

\usepackage{cite}
\usepackage{amsmath,amssymb,amsfonts}
\usepackage{algorithmic}
\usepackage{graphicx}
\usepackage{textcomp}
\usepackage{xcolor}
\usepackage{booktabs}
\usepackage{multirow}
\usepackage{tabularx}
\usepackage{array}
\usepackage{booktabs}
\usepackage{makecell}
\usepackage{url}
\usepackage[hidelinks]{hyperref}
\begin{document}

\title{TurnNat: Automatic Evaluation of Turn-Taking Naturalness in Dyadic Spoken Dialogue}

\author{
\IEEEauthorblockN{
Hao Zhang\textsuperscript{1,$\dagger$},
Thomas Thebaud\textsuperscript{1},
Georgi Tinchev\textsuperscript{2},
Venkatesh Ravichandran\textsuperscript{3},
Laureano Moro-Velázquez\textsuperscript{1,$\dagger$}
}
\IEEEauthorblockA{
\textsuperscript{1}Center for Language and Speech Processing, Johns Hopkins University, USA\\
\textsuperscript{2}Amazon Research, UK\\
\textsuperscript{3}Amazon, USA\\
hzhan276@jh.edu, laureano@jhu.edu
}
\thanks{
$\dagger$ Corresponding author.
}
}

\maketitle

\begingroup
\renewcommand{\thefootnote}{}
\footnotetext{\textbf{Code Resource}:
\url{https://github.com/TedZhangHao/turn-taking-naturalness}.}
\endgroup
\begin{abstract}

Turn-taking naturalness is central to full-duplex spoken dialogue systems, yet its automatic evaluation remains limited. Existing evaluations often rely on human judgments or behavior-specific timing metrics, making it difficult to compare heterogeneous timing failures within a unified framework. We propose TurnNat, a likelihood-based framework for automatic turn-taking naturalness evaluation in two-channel spoken dialogue. A causal turn-taking prediction model trained on natural conversations estimates future two-speaker voice-activity states, and the negative log-likelihood (NLL) of the observed future activity measures timing atypicality. TurnNat pools frame-level NLLs over turn-taking boundary units (TBUs) extracted from utterance onsets and offsets, and aggregates mean and tail TBU scores into a dialogue-level naturalness score. We further construct a controlled perturbation benchmark of paired natural and
perturbed dialogue clips, validated by human naturalness judgments. Experiments
on this benchmark show that TurnNat successfully identifies unnatural
turn-taking perturbations across heterogeneous timing failures.

\end{abstract}


\section{Introduction}

Natural spoken dialogue relies on fluent turn-taking and appropriate conversational timing. Human conversations exhibit remarkably precise temporal coordination, with turn transitions often occurring after only a few hundred milliseconds of silence. Such coordination requires participants to continuously interpret conversational cues, anticipate upcoming turn completions, and decide whether to hold, yield, backchannel, or take the conversational floor. In contrast, many spoken dialogue systems still rely on heuristic turn-taking strategies, such as fixed silence thresholds, often resulting in delayed responses or interruptions that reduce conversational naturalness \cite{castillo2025survey,sakuma2022response}.

As spoken dialogue systems move from command-style interfaces toward real-time speech-to-speech interaction, their success increasingly depends on how naturally they participate in the temporal flow of conversation \cite{ji2024wavchat,defossez2024moshi,fang2025llama,xie2024miniomni,wang2025freeze}. Recent full-duplex speech models aim to support low-latency responses, overlapping speech, or interruptions rather than treating dialogue as a sequence of isolated turns \cite{defossez2024moshi,lin2025fullduplexbench}. These interactional behaviors are not merely surface-level timing details: response latency affects perceived responsiveness and conversational naturalness \cite{maslych2025mitigating,roddy2020neural}, while the ability to handle interruptions or backchannels can influence users' trust and willingness to rely on voice assistants \cite{liu2025toward,baughan2023mixed}. Such properties are especially important for dialogue systems in settings such as healthcare or customer service, where user engagement and trust are central to effective interaction \cite{cevasco2024patient,sanjeewa2024empathic}. 

Recent benchmarks evaluate interactive spoken dialogue behaviors such as pause handling, backchanneling, turn-taking, and interruption management \cite{lin2025fullduplexbench,arora2025talking}. These efforts provide valuable diagnostics for spoken dialogue models, but their metrics are typically tied to specific interaction tasks or behavior categories. This motivates a complementary form of evaluation: an automatic metric for turn-taking naturalness within a single scoring framework, rather than evaluating each interactional behavior with a separate task-specific metric.
Turn-taking prediction models provide a natural starting point for this form
of evaluation, as they estimate upcoming conversational behavior from dialogue
context, including turn shifts, pauses, overlap, and backchannels
\cite{ekstedt2020turngpt,skantze2021turn,ekstedt2022vap}. Existing models use
different input signals, from linguistic context to speech and multimodal cues
\cite{lin2025predicting,wang2024turn}. Recent work such as DualTurn has explored dual-channel generative speech pretraining, where models learn conversational dynamics by predicting both speakers' future audio \cite{rajaa2026dualturn, wang2025ntpp}. 
Such models are useful for naturalness evaluation because they are trained to learn patterns of turn-taking dynamics. In particular, Voice Activity Projection's (VAP) formulation of turn-taking as future two-speaker voice-activity prediction provides a direct probabilistic target for scoring conversational timing \cite{ekstedt2022vap}. 

\vspace{-3pt}
We propose TurnNat, an automatic metric that assigns a dialogue-level
turn-taking naturalness score to two-channel spoken dialogue. TurnNat uses a causal turn-taking prediction model trained only on natural
conversations. The model defines a distribution over future two-speaker
voice-activity states, which TurnNat uses to assess whether observed speaker
activity patterns are typical of natural conversations. Locally unnatural timing patterns make the observed future activity less likely under this distribution. We also construct a
human-validated paired perturbation benchmark and show that TurnNat
distinguishes natural from perturbed timing patterns across heterogeneous
turn-taking failures.
Our contributions are threefold:

\begin{itemize}
\item We propose TurnNat, a unified likelihood-based framework for automatic
turn-taking naturalness evaluation, using the future two-speaker voice-activity
likelihood.

\item We construct a human-validated turn-taking perturbation benchmark covering
five localized turn-taking perturbations in natural human-to-human dyadic spoken
dialogue.

\item We instantiate the framework with VAP and DualTurn-based predictors,
showing that future-activity likelihood successfully distinguishes natural from
perturbed timing patterns across heterogeneous turn-taking failures.

\end{itemize}

\section{Related Work}

\subsection{Turn-Taking Prediction}

Turn-taking prediction provides the modeling basis for TurnNat. Skantze
\cite{skantze2021turn} reviews turn-taking as a continuous interactional process
involving shifts, holds, pauses, overlap, interruptions, and backchannels.
Earlier computational work models local turn-taking decisions, such as whether a
pause should lead to a hold or a speaker shift
\cite{schlangen2006reaction,meena2014data,johansson2015opportunities}. Skantze
\cite{skantze2017towards} further formulates turn-taking as continuous future
speech-activity prediction. TurnGPT predicts turn shifts from linguistic context
\cite{ekstedt2020turngpt}. VAP predicts future two-speaker voice activity from
speech and provides the probabilistic prediction target most directly related to
our scorer \cite{ekstedt2022vap}. Subsequent VAP-style work extends this target
to multilingual, prompt-guided, real-time, and multimodal settings
\cite{inoue2024multilingual,inoue2025prompt,inoue2024realtime,
russell2025visual}. Other recent models combine acoustic and language-model
features for turn-taking and backchannel prediction \cite{lin2025predicting,
wang2024turn}. DualTurn learns dyadic conversational dynamics through
dual-channel generative speech pretraining \cite{rajaa2026dualturn}. Taken together, these studies show that turn-taking prediction models can learn temporally structured expectations about two-speaker conversational behavior. Our work repurposes these models for evaluation, using future-activity likelihood as an automatic signal for turn-taking naturalness.
\vspace{-3pt}
\subsection{Evaluation of Turn-Taking Naturalness}

Prior evaluation of conversational timing often relies on human listening tests
or user studies. Roddy and Harte \cite{roddy2020neural} model dialogue response
timing and show that perceived timing naturalness depends on dialogue context.
Full-Duplex-Bench evaluates full-duplex spoken dialogue systems with
behavior-specific diagnostics such as pause handling, backchanneling, smooth
turn-taking, and interruption management \cite{lin2025fullduplexbench}. Its
follow-up benchmark focuses on overlap-heavy scenarios such as interruptions,
backchannels, side conversations, and ambient speech \cite{lin2026full}. Talking
Turns trains a supervised event-decision judge and uses event-specific
thresholds to evaluate whether a system speaks, continues, backchannels,
interrupts, or yields at appropriate moments \cite{arora2025talking}. These
benchmarks provide useful behavioral diagnostics, but they evaluate separate
interactional behaviors or event decisions. TurnNat instead scores heterogeneous
timing failures in a shared continuous space using the future two-speaker
voice-activity likelihood.
\section{Proposed TurnNat}
\begin{figure*}
    \centering
    \includegraphics[width=0.83\linewidth]{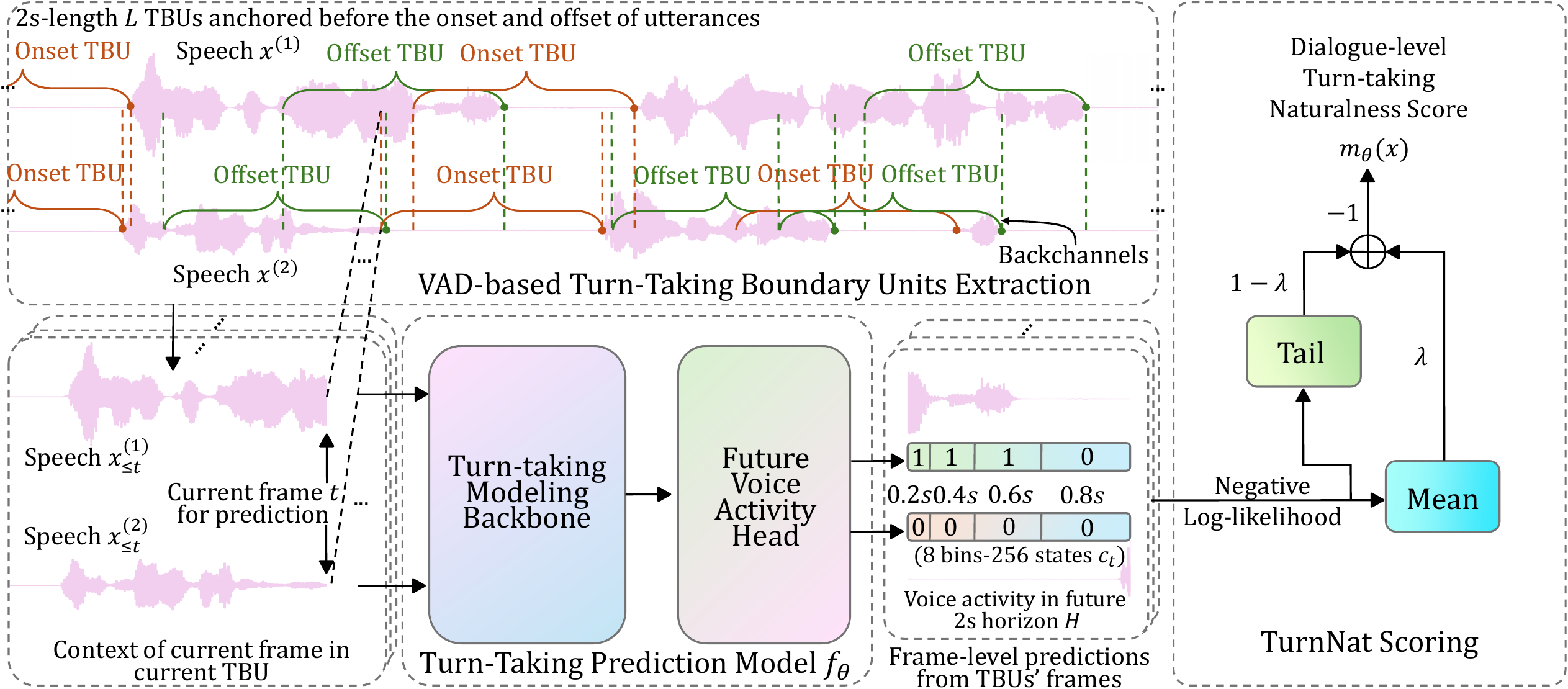}
    \caption{Overview of the TurnNat framework. TurnNat first extracts VAD-based
turn-taking boundary units from the two-channel dialogue, then uses a causal
turn-taking prediction model to assign likelihoods to future two-speaker
voice-activity states at frames inside these units. }
    \label{fig:TurnNat}
    \vspace{-12pt}
\end{figure*}

\subsection{Problem Formulation}

We study automatic evaluation of turn-taking naturalness in two-channel spoken
dialogue. Let \(x=(x^{(1)},x^{(2)})\) denote a dialogue segment, where
\(x^{(1)}\) and \(x^{(2)}\) correspond to the audio channels of speakers
1 and 2, respectively.

Let \(f_\theta\) denote a turn-taking prediction model with parameters
\(\theta\). For each frame \(t\), the model uses the available dialogue context
\(x_{\leq t}\) to estimate the likelihood of the observed future two-speaker
voice-activity state \(c_t\). TurnNat converts this likelihood into
frame-level negative log-likelihood values and aggregates them into a
dialogue-level turn-taking naturalness score
\(m_\theta(x)\in\mathbb{R}\), where higher values indicate more natural
turn-taking. TurnNat does not require human judgments, perturbation labels, or
manually annotated turn-taking events at inference time. Fig.~\ref{fig:TurnNat} shows the overall TurnNat framework in detail.
\vspace{-3pt}
\subsection{Turn-Taking Boundary Units (TBUs)}

TurnNat extracts TBUs from a cleaned two-speaker VAD sequence. We first identify
speech islands for each speaker and merge adjacent islands from the same speaker
when the gap is at most 1.0s, and the other speaker's active ratio in the gap is
at most 0.2. After merging, islands shorter than 200ms are discarded to remove
spurious fragments while retaining brief feedback events.

For each retained utterance candidate, we define two TBUs: one associated with the utterance onset and one associated with the utterance offset. For a boundary time $\tau_j$, the corresponding TBU \(u_j\) contains frames in the pre-boundary interval $[\tau_j - L, \tau_j]$, where $L = 2\mathrm{s}$. 
For a dialogue segment \(x\), we extract all TBUs
$\mathcal{U}(x)=\{u_j\}_{j=1}^{J}$, where ${J}$ is the total amount. Let \(\mathcal{T}(u_j)\) denote the set
of frames contained in unit \(u_j\). We define the set of all TBU
frames as follows:
\begin{equation}
\mathcal{T}_{\mathrm{TBU}}(x)
=
\bigcup_{u_j\in\mathcal{U}(x)} \mathcal{T}(u_j).
\end{equation}

\vspace{-4pt}
Although the scoring frames are selected before the boundary, each frame is later evaluated through a two-speaker voice-activity prediction over a future horizon. Thus, a TBU covers the local turn-taking region around an onset or offset boundary, including possible responses, gaps, overlaps, holds, or backchannels following the boundary.
\vspace{-3pt}
\subsection{Future Voice-Activity Prediction}
\noindent\textbf{Future Voice-Activity Prediction Target.}
TurnNat uses a causal future voice-activity prediction model as its likelihood source. Following VAP~\cite{ekstedt2022vap}, for each frame $t$, the model predicts both speakers' voice activity over a future horizon of $H=2\mathrm{s}$, divided into $K=4$ non-uniform bins: \([0,200]\), \([200,600]\), \([600,1200]\), and \([1200,2000]\) $\mathrm{ms}$.
This non-uniform discretization assigns a finer temporal resolution to near-future activity and a coarser resolution to farther-future activity, reflecting the increasing uncertainty of longer-horizon prediction while keeping the joint two-speaker future-activity state space tractable. This is well-suited to turn-taking evaluation, where local timing differences around upcoming speech activity are often perceptually important.
Let
\begin{equation}
\mathbf{b}_t =
\left[
b^{(1)}_{t,1},\ldots,b^{(1)}_{t,K},
b^{(2)}_{t,1},\ldots,b^{(2)}_{t,K}
\right]
\in \{0,1\}^{2K}
\end{equation}

denote the future voice-activity pattern after frame \(t\).
For speaker \(s\) and future bin \(k\), \(b^{(s)}_{t,k}=1\) if more than \(50\%\) of the frames in that bin are active according to the VAD, and \(b^{(s)}_{t,k}=0\) otherwise.
Each possible binary pattern $\mathbf{b}_t$ is indexed as one categorical future
voice-activity state:
\vspace{-4pt}
\begin{equation}
c_t = \mathrm{index}({\mathbf{b}_t}), \quad c_t \in \{1,\ldots,2^{2K}\}.     
\end{equation}

With \(K=4\), this yields \(2^8=256\) possible joint future voice-activity
states.

\noindent\textbf{Turn-Taking Prediction Model.}
The causal prediction model \(f_\theta\) consists of a turn-taking modeling backbone followed by a 256-way future voice-activity head (Fig.~\ref{fig:TurnNat}). Given the available dialogue context \(x_{\leq t}\), the model outputs a probability distribution $\mathbf{p}_t$ over the 256 joint future-activity states:
\begin{equation}
\mathbf{p}_\theta(t;x)
=
\mathrm{Softmax}
\left(
f_\theta(x_{\leq t})
\right),
\quad
\mathbf{p}_\theta(t;x) \in [0,1]^{256}.
\end{equation}
Here, \(\theta\) denotes the parameters of the turn-taking backbone and the
future voice-activity head. The specific backbone instantiations are described in
Section~\ref{sec:predictive_models}.




\noindent\textbf{Future-Activity Likelihood Training Objective.}
The turn-taking prediction models are trained only on natural human-to-human
dialogues (Section~\ref{sec:data}). Let \(\mathcal{D}_{\mathrm{nat}}\) denote
the natural training set. We minimize the weighted future-activity negative
log-likelihood over prediction frames:
\begin{equation}
\mathcal{L}_{\mathrm{train}}(\theta)
=
-
\frac{
\sum_{x\in\mathcal{D}_{\mathrm{nat}}}
\sum_{t\in\mathcal{T}_{\mathrm{pred}}(x)}
w_t(x)\log \mathbf{p}_{\theta}(t;x)[c_t]
}{
\sum_{x\in\mathcal{D}_{\mathrm{nat}}}
\sum_{t\in\mathcal{T}_{\mathrm{pred}}(x)}
w_t(x)
},
\end{equation}
where \(\mathcal{T}_{\mathrm{pred}}(x)\) is the prediction-frame set for
segment \(x\), and
\begin{equation}
w_t(x)=
\begin{cases}
\alpha, & t\in \mathcal{T}_{\mathrm{TBU}}(x),\\
1, & \text{otherwise}.
\end{cases}    
\end{equation}

Setting \(\alpha=1\) gives the standard uniformly weighted objective, while
\(\alpha>1\) emphasizes frames inside TBUs.





\subsection{TurnNat Scoring}

As depicted in Fig.~\ref{fig:TurnNat},  TurnNat converts the future-activity likelihood into frame-level NLL during evaluation time:
\begin{equation}
    \ell_\theta(t;x)=-\log \mathbf{p}_\theta(t;x)[c_t].
\end{equation}

A larger \(\ell_\theta(t;x)\) indicates that the observed future two-speaker activity is less typical under natural conversational dynamics.

For each TBU \(u_j\), TurnNat computes the unit-level NLL:
\vspace{-1pt}
\begin{equation}
s_\theta(u_j)
=
\frac{1}{|\mathcal{T}(u_j)|}
\sum_{t\in\mathcal{T}(u_j)}
\ell_\theta(t;x).
\end{equation}
Given the \(J\) TBU scores \(\{s_\theta(u_j)\}_{j=1}^{J}\), TurnNat aggregates
them using mean and tail terms:
\vspace{-5pt}
\begin{equation}
\begin{gathered}
\mathrm{MeanNLL}_\theta(x)
=
\frac{1}{J}
\sum_{j=1}^{J}
s_\theta(u_j),
\\
\mathrm{TailNLL}_\theta(x)
=
\mathrm{AvgTop}_{\rho}(\{s_\theta(u_j)\}_{j=1}^{J}),
\end{gathered}
\end{equation}
where \(\mathrm{AvgTop}_{\rho}(\cdot)\) averages the top fraction of highest-NLL TBUs,
so that strongly unnatural local turn-taking events are not diluted by the
global mean.  We use \(\rho=0.25\) in all
experiments. The final dialogue-level naturalness score is:
\begin{equation}
m_\theta(x)
=
-\left[
\lambda \mathrm{MeanNLL}_\theta(x)
+
(1-\lambda)\mathrm{TailNLL}_\theta(x)
\right],
\end{equation}
where \(\lambda=0.5\), giving equal weight to mean and tail
evidence. The negative sign converts NLL-oriented atypicality into naturalness, so higher \(m_\theta(x)\) indicates more natural turn-taking.
\section{Paired Perturbation Benchmark}
\label{sec:benchmark}

To evaluate turn-taking naturalness in a controlled setting, we construct a
paired perturbation benchmark from natural two-channel human-to-human dialogue.
Each benchmark item consists of a natural dialogue clip and a locally perturbed
counterpart derived from the same conversation. The paired design preserves
speaker identity, lexical content, recording condition, and most surrounding
dialogue context, so that differences between the two clips primarily reflect a
targeted change in turn-taking behavior. This allows automatic scorers to be
evaluated by whether they assign higher naturalness to the original clip than
to its perturbed counterpart.
\vspace{-7pt}
\subsection{Source Corpus and Splits}
\label{sec:data}

We use English two-channel dialogue recordings from the naturalistic portion of
the Seamless Interaction dataset~\cite{agrawal2025seamless}. Following the
released train, development, and test partitions, we select controlled subsets
and focus on ordinary prompted dyadic conversations based on the Interpersonal
Circumplex framework. We exclude task-oriented interaction types, such as
collaborative storytelling, or charades, because their
turn-taking patterns are more strongly constrained by the task format than by
open-ended conversational dynamics.

Table~\ref{tab:data_splits} summarizes the resulting speaker-disjoint splits.
The training split is used only to train future-activity prediction models, the
development split is used for model selection, and the held-out test split is
used as the source pool for constructing the paired perturbation benchmark.

\begin{table}[t]
\centering
\caption{Speaker-disjoint natural dialogue splits.}
\label{tab:data_splits}
\small
\setlength{\tabcolsep}{4.5pt}
\renewcommand{\arraystretch}{1.08}
\resizebox{\columnwidth}{!}{
\begin{tabular}{lrrrrl}
\toprule
\textbf{Split}
& \textbf{\# Dyads}
& \textbf{\# Speakers}
& \textbf{Hours}
& \textbf{Mean Dur.}
& \textbf{Usage} \\
\midrule
Train & 4,263 & 1,140 & 250.18 & 211.27s & Model training \\
Dev & 345 & 49 & 20.44 & 213.26s & Model selection \\
Test & 2,251 & 287 & 129.32 & 206.82s & Benchmark source \\
\bottomrule
\end{tabular}}
\vspace{-12pt}
\end{table}
\vspace{-5pt}
\subsection{Benchmark Construction}
\vspace{-3pt}
We sample 20--25s dialogue clips from the held-out test split. Candidate
regions are identified using the Silero voice activity detector \cite{silero_vad_2024}, with transcript text used
only for auditing and backchannel identification. Each crop is centered around a
turn-taking event targeted by one of the perturbation types, and its start and
end points are selected from regions where both speakers are silent to avoid
truncating ongoing speech. After basic audio and VAD quality control, we randomly downsample 200
natural--perturbed pairs per perturbation type, yielding 1,000 pairs for
automatic evaluation.

The perturbation magnitudes are chosen to be clearly outside typical
human--human response timing while remaining plausible for listening
evaluation. Prior work reports that many human responses begin within a few
hundred milliseconds of the previous turn ending, for example, from about
-200 to 400ms~\cite{sakuma2022response}. We therefore use larger timing shifts
to create controlled degradations rather than borderline timing variation.

For hold and shift perturbations, we select clean candidate regions using the
VAP shift/hold criterion~\cite{ekstedt2022vap}: a mutual-silence interval is
bounded by a single active speaker in a 1s pre-offset region and a single active
speaker in a 1s post-onset region. If the active speaker is before and after the
silence differs, the region is treated as a speaker shift; if the speaker is the same, it is treated as a speaker hold. Backchannels are selected using the
transcript-based criterion of Wang et al.~\cite{wang2024turn}: isolated one- or
two-word listener responses, such as ``yeah'' or ``mmhmm'', during the other
speaker's turn.

The benchmark covers five localized timing and floor-management perturbations as follows:

\begin{itemize}
\item \textbf{Late response}: We delay a responding utterance by 1.2--2.0s,
creating an abnormally long response gap.

\item \textbf{Early entry}: We advance a responding utterance by 1.2--2.5s,
creating premature overlap with the current speaker.

\item \textbf{Hold instead of shift}: We start from a clean speaker-shift
region and remove the responding turn, so the original speaker appears to
retain the floor instead of shifting it.

\item \textbf{Shift instead of hold}: We start from a clean speaker-hold region
and insert a complete turn from the other speaker, so the dialogue appears to
shift speakers where the original speaker would normally continue.

\item \textbf{Excessive backchanneling}: We insert two to three additional
listener-feedback events into regions where the listener is originally silent,
using naturally occurring backchannels from the same speaker and conversation.
\vspace{-12pt}
\end{itemize}
\vspace{-5pt}
\subsection{Human Validation Protocol}
\vspace{-3pt}
Human judgments are collected to assess the perceptual validity of the
perturbation benchmark. For validation, we sample a balanced subset of 150
natural--perturbed pairs, with 30 pairs from each perturbation type. Each pair
is rated by three independent native English-speaking annotators.

In each trial, annotators listen to the two clips in randomized order, choose
which clip has more natural turn-taking (A/B test), and rate each clip on a five-point scale
of naturalness and artifact scales. The resulting judgments are analyzed using the
human evaluation metrics described in Section~\ref{sec:human_eval_metrics}. 
\section{Experiments}


The turn-taking prediction models are trained on the natural training split and selected on
the development split described in Section~\ref{sec:data}. All automatic
evaluations are conducted on the paired perturbation benchmark constructed from
the held-out test split, as described in Section~\ref{sec:benchmark}. Human judgments
 are used only to validate the benchmark.



\vspace{-3pt}
\subsection{Human Evaluation Metrics}
\label{sec:human_eval_metrics}

For human validation, we report two A/B preference statistics at different
aggregation levels. Natural preference rate is computed over individual
annotations, as the fraction of judgments choosing the original non-perturbed
clip as having more natural turn-taking. Majority-natural pair rate is computed
after majority voting within each pair, as the fraction of pairs where the
natural clip wins the majority vote.

For five-point ratings, we report the mean naturalness-rating difference
between natural and perturbed clips, where positive values favor natural clips.
We also report mean artifact ratings and the perturbed-minus-natural artifact
difference to check whether audible artifacts confound naturalness judgments.
\vspace{-3pt}
\subsection{Prediction Model Comparison}
\label{sec:predictive_models}

We compare TurnNat scorers instantiated with VAP and DualTurn backbones. The comparison includes released checkpoints and fully fine-tuned variants, using
each backbone's native future-activity target or the shared 256-way
categorical target used by TurnNat. 

\noindent\textbf{VAP.}
VAP uses a CPC speech encoder, a causal Transformer, and a 256-way
classification head that predicts the joint future voice activity of both
speakers over a 2s horizon~\cite{ekstedt2022vap}. We evaluate the released VAP
checkpoint directly and a fully fine-tuned VAP variant trained with the same
256-way categorical future-activity target.

\noindent\textbf{DualTurn.}
DualTurn learns causal representations through dual-channel
generative speech pretraining with a frozen Mimi encoder and a Qwen
backbone~\cite{rajaa2026dualturn}. Its original training includes multiple
turn-taking signals, including VAD, future activity, end-of-turn, hold,
beginning-of-turn, and backchannel prediction. The Mimi encoder is frozen in all experiments. 

For comparison with the 256-way categorical TurnNat scorer, we evaluate two
DualTurn scoring forms. The first uses DualTurn's native eight Bernoulli future-activity targets, defined
by 4 horizon bins per speaker at 0.24, 0.48, 0.96, and 2.00\,s. The
second replaces this output with the shared 256-way categorical head used by
TurnNat.

\noindent\textbf{Training Details.}
We train each model for up to 5 epochs with AdamW and a batch size of
8, and select the checkpoint with the lowest development \textbf{future-activity loss}.
Early stopping is used to reduce overfitting. Every experiment is conducted on a single NVIDIA A100 80GB GPU. 
\vspace{-3pt}

\subsection{Automatic Evaluation Metrics}

For each natural--perturbed pair
\((x_i^{\mathrm{nat}}, x_i^{\mathrm{pert}})\), an automatic metric produces
dialogue-level naturalness scores \(m_\theta(x_i^{\mathrm{nat}})\) and
\(m_\theta(x_i^{\mathrm{pert}})\), where higher values indicate more natural
turn-taking. We define the paired score difference as
\begin{equation}
\Delta m_{\theta,i} =
m_\theta(x_i^{\mathrm{nat}}) -
m_\theta(x_i^{\mathrm{pert}}).
\end{equation}
A positive \(\Delta m_{\theta,i}\) indicates that the natural clip receives a
higher naturalness score than its perturbed counterpart.

\noindent\textbf{Concordance Index.}
The concordance index (C-index) measures global separability between natural and
perturbed clips:
\begin{equation}
\mathrm{C\mbox{-}index}
=
\frac{
\sum_{i=1}^{N}
\sum_{j=1}^{N}
\mathbb{I}
\left[
m_\theta(x_i^{\mathrm{nat}}) > m_\theta(x_j^{\mathrm{pert}})
\right]
}{
\sum_{i=1}^{N}
\sum_{j=1}^{N}
\mathbb{I}
\left[
m_\theta(x_i^{\mathrm{nat}}) \neq m_\theta(x_j^{\mathrm{pert}})
\right]
}.
\end{equation}

\noindent\textbf{Pairwise Accuracy.}
Pairwise accuracy measures whether the metric ranks each matched pair in the
expected direction:
\begin{equation}
\mathrm{Acc}_{\mathrm{pair}}
=
\frac{1}{N}
\sum_{i=1}^{N}
\mathbb{I}
\left[
\Delta m_{\theta,i} > 0
\right].
\end{equation}
We report the mean \(\Delta m_\theta\) as a within-scorer separation diagnostic;
C-index and matched-pair accuracy are the primary scale-free comparison metrics.




\section{Results}
We first verify that the constructed turn-taking benchmark produces
human-perceived differences between natural and perturbed pairs. We then evaluate whether TurnNat
distinguishes natural from perturbed clips on the same benchmark.
\vspace{-2pt}
\vspace{-10pt}
\subsection{Human Validation of the Perturbation Benchmark}
\begin{table}[t]
\centering
\caption{Human validation results by perturbation type.}
\label{tab:human_eval_by_type}
\footnotesize
\setlength{\tabcolsep}{6pt}
\renewcommand{\arraystretch}{1.08}
\begin{tabular}{lcccc}
\toprule
Type &
Nat. pref. $\uparrow$ &
Maj. nat. $\uparrow$ &
$\Delta$Nat. $\uparrow$ &
$\Delta$Art. $\to 0$ \\
\midrule
Late Response  & 0.756 & 0.867 & 0.667 & -0.111 \\
Early Entry    & 0.678 & 0.700 & 0.700 &  0.667 \\
Hold Instead   & 0.733 & 0.867 & 0.900 &  0.500 \\
Shift Instead  & 0.656 & 0.700 & 0.533 &  0.389 \\
Excessive BC   & 0.678 & 0.767 & 0.589 &  0.000 \\
\midrule
Overall        & 0.700 & 0.780 & 0.678 &  0.289 \\
\bottomrule
\end{tabular}

\vspace{2pt}
\begin{minipage}{0.98\columnwidth}
\footnotesize
\emph{Note.} Nat. pref. is the fraction of judgments choosing the natural clip.
Maj. nat. is the fraction of pairs with majority-natural preference.
$\Delta$Nat. is natural minus perturbed naturalness rating (1--5).
$\Delta$Art. is perturbed minus natural artifact rating (1--5), where values
closer to 0 indicate better artifact control.
\end{minipage}
\vspace{-12pt}
\end{table}

As shown in Table~\ref{tab:human_eval_by_type}, human judgments consistently
favor natural turn-taking. Annotators prefer the natural clip in 70.0\%
(95\% Wilson CI: 65.6--74.1) of individual pairwise judgments. After majority
voting within each pair, the natural clip is preferred in 78.0\% of pairs
(95\% Wilson CI: 70.7--83.9). Natural clips also receive higher five-point
naturalness ratings by 0.678 points on average. A participant-level
paired bootstrap confirms that the naturalness-rating difference is significantly larger than the artifact-rating difference by 0.389 points (95\% CI: [0.138, 0.651]),
suggesting that the observed preference pattern is not primarily driven by
waveform artifacts.

\begin{figure*}[t]
\centering

\includegraphics[width=0.93\linewidth]{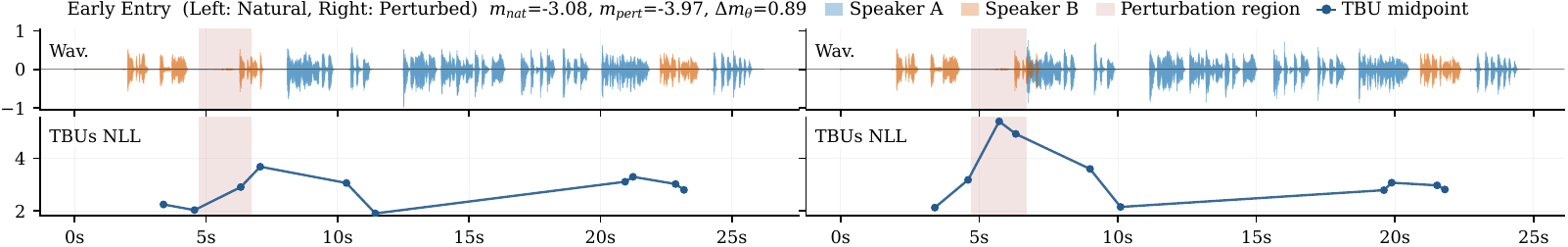}\\
\includegraphics[width=0.93\linewidth]{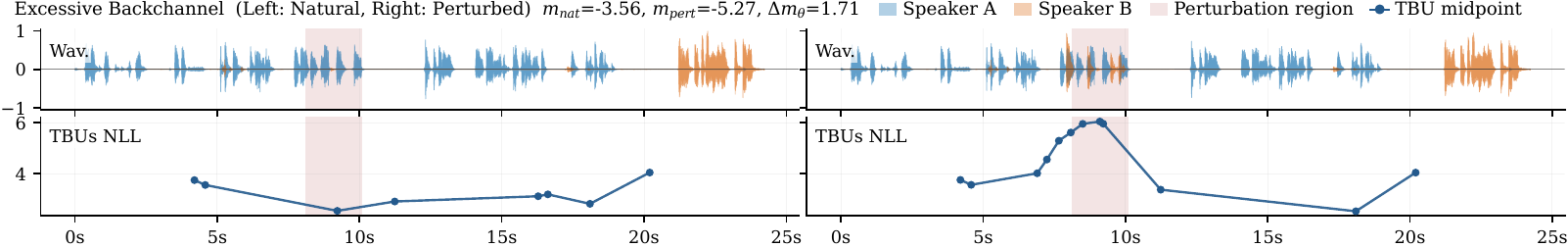}
\caption{Representative natural--perturbed pairs. Each row shows one perturbation type, with the
    natural clip on the left and the perturbed clip on the right. Shaded regions
    indicate the perturbation region, and markers show TBU midpoints. Perturbed
    clips show localized increases in TBUs NLL.}
\label{fig:qualitative_examples}
\vspace{-11pt}
\end{figure*}
\vspace{-6pt}
\begin{table*}[t]
\centering
\caption{Automatic discrimination results for dialogue-level TurnNat scores evaluated on the perturbation benchmark.}
\label{tab:auto_discrimination}
\setlength{\tabcolsep}{3.0pt}
\begin{tabular}{llllc|ccc|cc|cc|cc|cc|cc}
\toprule
\multicolumn{3}{c}{Architecture} &
\multicolumn{2}{c}{Training} &
\multicolumn{3}{c}{Overall} &
\multicolumn{10}{c}{By perturbation type: C-index $\mid$ $\mathrm{Acc}_{\mathrm{pair}}$ (\%) $\uparrow$} \\
\cmidrule(lr){1-3}
\cmidrule(lr){4-5}
\cmidrule(lr){6-8}
\cmidrule(lr){9-18}
ID & Backbone & Output &
Adapt. & $\alpha$ &
$\Delta m_\theta$ $\uparrow$ & C-index $\uparrow$ & $\mathrm{Acc}_{\mathrm{pair}}$ (\%) $\uparrow$ &
\multicolumn{2}{c}{Late resp.} &
\multicolumn{2}{c}{Early entry} &
\multicolumn{2}{c}{Hold instead} &
\multicolumn{2}{c}{Shift instead} &
\multicolumn{2}{c}{Excess BC} \\
\midrule
V0 & VAP & 256-way Cat. & None &
-- & \textbf{0.60} & 0.633 & 80.6 &
.562 & 90.0 & .671 & 91.0 & .634 & 82.0 & .547 & 66.0 & .771 & 74.0 \\
D0 & DualTurn & indep. Bern. & None &
-- & 0.47 & 0.645 & 77.5 &
.535 & 66.0 & .675 & 85.0 & .653 & 73.0 & .600 & 83.0 & .763 & 80.5 \\
V1 & VAP & 256-way Cat. & FT &
1 & 0.36 & 0.641 & 80.2 &
.571 & 79.5 & .672 & 85.0 & .631 & 74.0 & .567 & 82.5 & .767 & 80.0 \\
D1 & DualTurn & indep. Bern. & FT &
1 & 0.47 & 0.663 & 81.2 &
.576 & 75.5 & .672 & 82.0 & .740 & 73.0 & .583 & 91.5 & .769 & 84.0 \\
D2 & DualTurn & indep. Bern. + Aux. & FT &
1 & 0.40 & 0.657 & 81.5 &
.565 & 78.0 & .681 & 86.0 & .719 & 71.5 & .576 & \textbf{92.0} & .766 & 80.0 \\
D3 & DualTurn & 256-way Cat. & FT &
1 & 0.44 & 0.660 & 83.3 &
.570 & 82.0 & \textbf{.718} & 90.0 & .592 & 78.0 & .621 & 82.0 & .790 & 84.5 \\
D4 & DualTurn & 256-way Cat. + Aux. & FT &
1 & 0.45 & 0.670 & 86.2 &
.607 & 93.5 & .712 & \textbf{93.5} & .630 & 80.0 & .602 & 79.5 & .795 & 84.5 \\
\midrule
D2 & DualTurn & indep. Bern. + Aux. & FT &
3 & 0.47 & 0.669 & 81.7 &
.575 & 78.5 & .682 & 81.5 & \textbf{.742} & 75.5 & .596 & \textbf{92.0} & .773 & 81.0 \\
D4 & DualTurn & 256-way Cat. + Aux. & FT &
3 & 0.46 & \textbf{0.676} & 87.3 &
\textbf{.619} & 94.0 & .715 & 92.0 & .633 & 83.5 & .606 & 82.0 & .810 & 85.0 \\
D4 & DualTurn & 256-way Cat. + Aux. & FT &
8 & 0.45 & \textbf{0.676} & \textbf{88.0} &
\textbf{.619} & \textbf{95.0} & .713 & 92.5 & .609 & \textbf{84.5} & \textbf{.628} & 81.0 & \textbf{.813} & \textbf{87.0} \\
\bottomrule
\end{tabular}

\vspace{2pt}
\begin{minipage}{0.96\linewidth}
\footnotesize
\raggedright

\emph{Note.} $\Delta m_\theta$ is the mean paired score difference
\(m_\theta(x^{nat})-m_\theta(x^{pert})\), used as a within-scorer separation
diagnostic. Each perturbation-type group reports C-index followed by $\mathrm{Acc}_{\mathrm{pair}}$.
$\alpha$ is the TBU training weight. Adapt. denotes adaptation setting; FT denotes full fine-tuning; ``indep.
Bern.'' denotes DualTurn's native Bernoulli future-activity targets. Aux. denotes auxiliary losses for the original DualTurn turn-taking signal heads.
\vspace{-15pt}
\end{minipage}
\end{table*}

\begin{table}[t]
\centering
\caption{Aggregation ablation using the D4 scorer with \(\alpha=8\).}
\label{tab:aggregation_ablation_full}
\setlength{\tabcolsep}{8pt}
\begin{tabular}{lccc}
\toprule
Aggregation &
$\Delta m_\theta$ $\uparrow$ &
C-index $\uparrow$ &
$\mathrm{Acc}_{\mathrm{pair}}$ (\%) $\uparrow$ \\
\midrule
All frames mean & 0.20 $\pm$ 0.01 & 0.609 & 84.7 \\
TBU mean only   & 0.37 $\pm$ 0.03 & 0.662 & 87.4 \\
TBU tail only   & \textbf{0.54 $\pm$ 0.05} & 0.674 & 84.5 \\
TBU mean + tail & 0.45 $\pm$ 0.04 & \textbf{0.676} & \textbf{88.0} \\
\bottomrule
\end{tabular}

\vspace{2pt}
\begin{minipage}{0.96\columnwidth}
\footnotesize
\emph{Note.} \(\Delta m_\theta\) is reported with 95\% bootstrap CI half-widths. $\mathrm{Acc}_{\mathrm{pair}}$ is matched-pair accuracy. All values use D4 with
\(\alpha=8\).
\vspace{-25pt}
\end{minipage}

\end{table}

\subsection{TurnNat Evaluation on the Perturbation Benchmark}

\noindent\textbf{Overall discrimination.}
Table~\ref{tab:auto_discrimination} reports automatic discrimination results on
the perturbation benchmark. We focus on matched-pair accuracy and C-index:
matched-pair accuracy tests whether a scorer assigns higher naturalness to a
natural clip than to its matched perturbed counterpart, while C-index measures
global natural--perturbed separability across clips. The best configuration,
D4 with $\alpha=8$, achieves 88.0\% matched-pair accuracy (95\% Wilson CI:
85.8--89.9) and a C-index of 0.676, improving over the VAP baselines
(V0: 80.6\%, V1: 80.2\%) by about 7--8 absolute points and over the unadapted
DualTurn Bernoulli scorer (D0: 77.5\%) by 10.5 points. The gap between
matched-pair accuracy and C-index suggests that paired comparisons are more
stable than global score comparisons across heterogeneous dialogue contexts.

\noindent\textbf{Architecture ablation.}
Per-type results show that D4 with $\alpha=8$ is strongest overall and performs
particularly well on late response, early entry, hold-instead, and excessive
backchanneling. The main exception is shift-instead perturbations, where the
Bernoulli auxiliary scorer reaches the highest matched-pair accuracy. This
suggests that independent horizon targets and categorical joint-state targets
emphasize different aspects of turn-taking behavior, while the categorical
auxiliary model provides the most balanced discrimination overall.

\noindent\textbf{Effect of TBU weighting.}
Emphasizing TBUs during training further improves the final score. Increasing
$\alpha$ from 1 to 3 and 8 improves D4 from 86.2\% to 87.3\% and 88.0\% pair
accuracy, respectively, while the C-index increases from 0.670 to 0.676. The two
weighted D4 variants are close, suggesting that TBU weighting is beneficial but
not highly sensitive within this range. Applying the same weighting to the
Bernoulli auxiliary scorer yields a small gain, from 81.5\% to 81.7\%,
suggesting that the strongest results require both the categorical
future-activity target and auxiliary DualTurn supervision.

\noindent\textbf{Perturbation-type analysis.}
Per-type results show that D4 with $\alpha=8$ is strongest overall and performs
well on late response, early entry, hold instead of shift, and excessive
backchanneling. The main exception is shift-instead perturbations, where the
Bernoulli auxiliary scorer reaches the highest matched-pair accuracy. This
suggests that independent horizon targets and categorical joint-state targets
emphasize different aspects of turn-taking behavior, while the categorical
auxiliary model provides the most balanced discrimination overall.

\noindent\textbf{Aggregation ablation.}
Table~\ref{tab:aggregation_ablation_full} shows that TBU-based aggregation
outperforms all-frame averaging. Combining mean and tail TBU pooling gives the
best overall result, reaching 88.0\% pair accuracy and a C-index of 0.676.

\noindent\textbf{Visualized qualitative case study.}
Fig.~\ref{fig:qualitative_examples} shows representative natural--perturbed
pairs for early entry and excessive backchanneling. Perturbed clips exhibit
localized TBU-NLL increases near the modified region, providing an interpretable
view of TurnNat's dialogue-level scores. Additional examples are provided in the
released repository.

\vspace{-4pt}


\section{Conclusion}

We introduced TurnNat, a likelihood-based framework for evaluating turn-taking
naturalness in dyadic spoken dialogue, together with a paired perturbation
benchmark whose turn-taking degradations are validated by human judgments. The
benchmark covers five localized timing and floor-management failures. TurnNat
places heterogeneous turn-taking behaviors in a shared likelihood-based
evaluation space, producing a single dialogue-level naturalness score while
still allowing local inspection through high-surprisal TBUs. Across VAP and
DualTurn-based instantiations, future two-speaker voice-activity prediction
provides likelihood distributions that reliably distinguish natural from
perturbed clips. These results support future-activity likelihood as a unified
signal for evaluating heterogeneous turn-taking failures.
\vspace{-5pt}
\section{Limitations}
\vspace{-2pt}
This work evaluates controlled perturbations of natural human-to-human
dialogue. While the paired design isolates turn-taking timing differences, it
does not cover all timing failures in deployed spoken dialogue systems, such as
system-side latency, semantic misunderstandings, or prosodic mismatch. TurnNat also focuses on future two-speaker voice activity
and may miss cases where naturalness depends on lexical content, discourse
intent, speaker relationship, or task-dependent naturalness cues. Future work would extend the
benchmark to real human--AI conversations, more languages, and broader domains.

\section{Generative AI Use Disclosure}
The authors used ChatGPT (OpenAI) to assist with grammar checking and proofreading the manuscript. These tools were not used to generate any scientific content, results, or conclusions. All authors reviewed and take full responsibility for the final content of this paper.
\bibliographystyle{IEEEtran}
\bibliography{IEEEabrv,IEEEmain}

\end{document}